\documentclass[compsoc]{IEEEtran}
\usepackage[nocompress]{cite}
\usepackage[pdftex]{graphicx}
\usepackage[cmex10]{amsmath}
\usepackage{amsthm, amssymb}
\usepackage{bm}
\usepackage{xfrac}
\usepackage{algorithmic}
\usepackage{array}
\usepackage{mdwmath}
\usepackage{mdwtab}
\usepackage{multirow}
\usepackage{dblfloatfix}
\usepackage{url}
\newtheorem{definition}{Definition}
\begin{document}
\title{GAP2WSS: A Genetic Algorithm based on the Pareto Principle for Web Service Selection}
\author{SayedHassan~Khatoonabadi, Shahriar~Lotfi, and Ayaz~Isazadeh \thanks{The authors are with the Department of Computer Science, Faculty of Mathematical Sciences, University of Tabriz, Tabriz, Iran. Email: s\_hkatoonabadi91@ms.tabrizu.ac.ir, \{shahriar\_lotfi, isazadeh\}@tabrizu.ac.ir.}}
\markboth{}{Khatoonabadi \MakeLowercase{\textit{et al.}}: A Genetic Algorithm based on the Pareto Principle for Web Service Selection}
\IEEEtitleabstractindextext{%
\begin{abstract}
Despite all the progress in Web service selection, the need for an approach with a better optimality and performance still remains. This paper presents a genetic algorithm by adopting the Pareto principle that is called GAP2WSS for selecting a Web service for each task of a composite Web service from a pool of candidate Web services. In contrast to the existing approaches, all global QoS constraints, interservice constraints, and transactional constraints are considered simultaneously. At first, all candidate Web services are scored and ranked per each task using the proposed mechanism. Then, the top 20 percent of the candidate Web services of each task are considered as the candidate Web services of the corresponding task to reduce the problem search space. Finally, the Web service selection problem is solved by focusing only on these 20 percent candidate Web services of each task using a genetic algorithm. Empirical studies demonstrate this approach leads to a higher efficiency and efficacy as compared with the case that all the candidate Web services are considered in solving the problem.
\end{abstract}
\begin{IEEEkeywords}
M.2.0.b Science of service composition, M.4.0.f Quality of service, M.5.2.g.1 Composition, M.10.0.a Business process modeling and management
\end{IEEEkeywords}}
\maketitle
\IEEEraisesectionheading{\section{Introduction}\label{sec:introduction}}
\IEEEPARstart{A}{Web} service is an enabling technology for realization of service-oriented architecture~\cite{erl2005service} and service-oriented computing~\cite{papazoglou2003service} paradigms. It provides a platform-independent framework by means of XML~\cite{ws-xml} standards, which enables services to be delivered over the Internet. Formally, a Web service~\cite{ws-gloss} is defined as a software system which supports interoperable machine-to-machine interactions over a network, and has an interface described in a machine-processable format (e.g., WSDL~\cite{ws-wsdl}). Meanwhile, interactions with a Web service are carried out by SOAP-messages~\cite{ws-soap0, ws-soap1, ws-soap2, ws-soap3} as described in its WSDL.
\par
Service providers develop different Web services and release them in service registries (e.g., UDDI~\cite{uddi}). The service registries maintain technical details of the Web services and information of their corresponding service provider, such as address and contact. Service consumers retrieve and utilize those information from the registry to make use of their demanded Web services.
\par
In most cases, a sole Web service provides a limited functionality which is not adequate to accomplish sophisticated tasks. For reasons like cost, time, and ease of implementation, many prefer to concentrate on their own core task, and outsource other tasks through reusing existing Web services. The process of seamless, flexible, and loosely-coupled integration of various Web services to fulfill a complex task is called Web service composition, and the resultant value-added Web service is called a composite Web service. A concrete example of a composite Web service is a travel planner comprised of a flight booking, a hotel booking, and a car rental Web service.
\subsection{Quality of Service (QoS)}
QoS, as the name suggests, defines the quality of a Web service, and includes its quantitative non-functional properties. This can consist of generic as well as application-specific attributes. In other words, QoS provides a way to distinguish between Web services with the same functionality.
\par
The W3C defined a number of generic QoS attributes in~\cite{ws-qos}. In addition, Al-Masri and Mahmoud~\cite{al2008investigating, al2007qos, al2007discovering} provided the QWS dataset (\url{www.uoguelph.ca/~qmahmoud/qws/}) which contains the measured values of nine QoS attributes for over 2500 actual Web services. A formal definition of these QoS attributes is presented below:
\begin{IEEEitemize}
\item
Response Time: is defined as the time taken to send a request to a Web service and receive a response from it, expressed in milliseconds.
\item
Availability: is defined as the ratio of the number of successful invocations to the total number of invocations in a Web service, expressed in percentage.
\item
Throughput: is defined as the total number of invocations per second in a Web service, expressed as invokes/second.
\item
Successability: is defined as the ratio of the number of responses to the number of requests in a Web service, expressed in percentage.
\item
Reliability: is defined as the ratio of the number of successful messages to the total number of messages in a Web service, expressed in percentage.
\item
Compliance: is defined as the extent to which a Web service's WSDL follows WSDL Specification~\cite{ws-wsdl}, expressed in percentage.
\item
Best Practices: is defined as the extent to which a Web service follows WS-I Basic Profile~\cite{wsi-basicprofile}, expressed in percentage.
\item
Latency: is defined as the time taken for a Web service to process a request, expressed in milliseconds.
\item
Documentation: is defined as the extent to which a Web service is documented in its WSDL by description tags, expressed in percentage.
\end{IEEEitemize}
\par
Different functionally equivalent Web services provide different trade-offs between the QoS attributes. In other words, a Web service may be superior to some other Web services in terms of some QoS attributes, but may not be as good as them in terms of some other QoS attributes. Accordingly, QoS plays a significant role in Web service composition.
\subsection{The Web Service Selection Problem}
\begin{figure}[!t]
\centering
\includegraphics[width=\columnwidth]{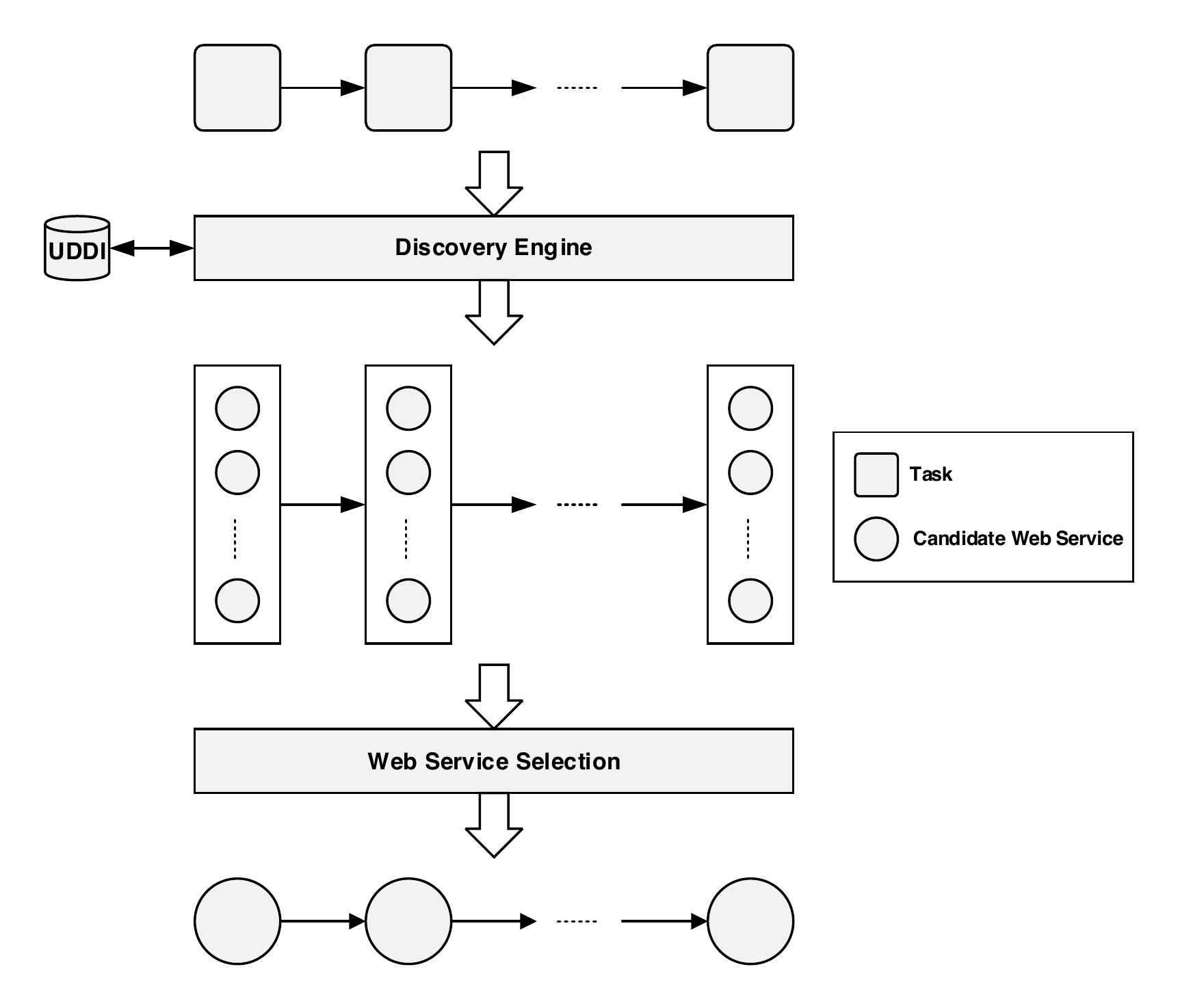}
\caption{A conceptual overview of the Web service selection problem}
\label{fig:problem}
\end{figure}

Given a composition request described in a workflow language (e.g., BPEL~\cite{ws-bpel}, YAWL~\cite{van2005yawl}), the discovery engine searches the registry to detect functional matching Web services for each task of the workflow using Web service descriptions. As a result, a set of functionally equivalent candidate Web services with different QoS values is acquired per each task. The objective of the Web service selection problem is to select one candidate Web service from each set, such that the QoS values of the yielded composite Web service become the best along all the possible combinations, while all the specified constraints are satisfied. This is also known as the QoS-aware Web service composition problem in the literature. A conceptual overview of the problem is shown in Fig.~\ref{fig:problem}.
\par
The Web service selection is a highly complicated and challenging problem. On the one hand, a composite Web service may be comprised of many tasks, and on the other hand, the number of available candidate Web services for each task is increasing. According to~\cite{al2008investigating}, there has been more than 130 percent growth in the number of available candidate Web services in the period between October 2006 to October 2007.  In addition, with the spread of cloud computing~\cite{armbrust2010view}, a sharp increase in the number of available candidate Web services is extremely anticipated.
\par
The Web service selection can be modeled as a multidimensional multiple-choice knapsack problem~\cite{kellerer2004knapsack} which is known to be NP-hard~\cite{arora2009computational} in the strong sense. In a multidimensional multiple-choice knapsack problem, different groups of items exist, where each item has a certain value. Groups, items, and the value of each item in this problem corresponds to tasks, candidate Web services, and the utility value of each candidate Web service in the Web service selection problem, respectively. Let $n$, $m$ represent the number of tasks and the number of candidate Web services for each task, respectively. Then, $m^n$ distinct combinations of them exist. Therefore, with the growth of problem size, the computational complexity increases exponentially. Combined with the constraints, the problem becomes even more complicated.
\par
The significance of the Web service selection problem originates from two points. Firstly, since the QoS values of a composite Web service strongly depend on the QoS values of the Web services which comprise it, the efficacy of the applied approach is of high importance. Secondly, the efficiency of the applied approach is crucial in real-time applications.
\par
Consequently, an ideal approach to solving the Web service selection problem shall
\begin{IEEEenumerate}
\item
require the minimum possible time to obtain the solution,
\item
obtain the best possible QoS values for the solution, and
\item
consider all the required constraints to obtain the solution.
\end{IEEEenumerate}
\par
Unfortunately, the performance and the optimality of the approach are in contrast practically. This means that obtaining the best solution in the minimum time is virtually impossible. Hereupon, the importance of performance and optimality totally depends on the application. For instance, some may prefer performance over optimality and some may prefer otherwise. Intuitively, it can be inferred that what matters generally is not the performance or the optimality of the approach separately, but is a reasonable balance between them.
\par
In addition to the performance and the optimality, an approach should also consider all the required constraints to enlarge its applicability. This means that the more constraints considered, the more usability the approach will have in real and enterprise applications.
\par
It should be noted that in cases which concern is only taken to the optimality of the solution, exact approaches such as integer linear programming~\cite{zeng2003quality, zeng2004qos, ardagna2007adaptive} and exhaustive search which obtain the optimal solution can be used. However, these approaches are not scalable and may face an ultra-high computational time with respect to the huge number of tasks and candidate Web services. In addition, other approaches such as~\cite{yu2007efficient, alrifai2009combining, alrifai2010selecting, alrifai2012hybrid} are proposed to increase the scalability which obtain the near-optimal solution. Still, these approaches are not sufficiently scalable.
\subsection{Contribution of the Paper}
The main contributions of this paper are stated as follows:
\begin{IEEEitemize}
\item
Formulated the Web service selection problem by considering global QoS constraints, interservice constraints, and transactional constraints.
\item
Proposed a more efficient and effective approach for solving the Web service selection problem using the Pareto principle and genetic algorithm.
\item
Designed a fitness function to support global QoS constraints, interservice constraints, and transactional constraints in genetic algorithm.
\item
Designed a mechanism to score and rank candidate Web services of different tasks to utilize the Pareto principle.
\end{IEEEitemize}
\subsection{Paper Outline}
The rest of the paper is organized as follows. The next section presents a brief survey of the previous works. Section~3 formulates the Web service selection problem. Section~4 elaborates GAP2WSS in details. Section~5 presents the experimental evaluations. Finally, section~6 concludes the paper and puts forward some future topics of research.
\section{Related Work}
The Web service selection problem is a non-trivial popular research topic which has received the attention of many researchers for over a decade. There are two general approaches for the Web service selection problem: evolutionary and non-evolutionary approaches. Typically, evolutionary approaches have higher efficiency and lower efficacy than non-evolutionary approaches. In the following, a brief survey of the previous works is presented.
\begin{figure*}[!t]
\centering
\includegraphics[width=\textwidth]{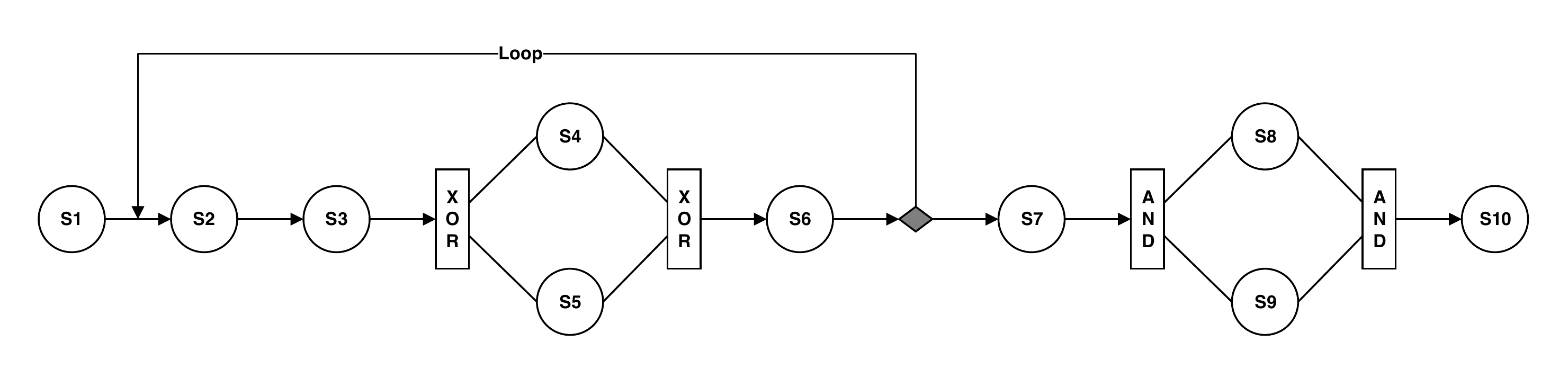}
\caption{The workflow of a composite Web service}
\label{fig:workflow}
\end{figure*}
\subsection{Evolutionary Approaches}
Ai and Tang~\cite{ai2008penalty, ai2008qos, tang2010hybrid} considered interservice constraints and utilized genetic algorithm~\cite{goldberg2006genetic} to obtain a near-optimal solution. They encoded each solution as an integer vector. In~\cite{ai2008penalty}, a penalty mechanism is used to penalize solutions which violate the constraints based on the degree of violation. In~\cite{ai2008qos}, a hill climbing repair mechanism is used to fix solutions which violate the constraints. In~\cite{tang2010hybrid}, a hybrid mechanism is used which penalizes solutions that violate the constraints based on the degree of violation, and optimizes all the solutions using a local optimizer. In~\cite{ai2008penalty, ai2008qos}, a single-point crossover and in~\cite{tang2010hybrid}, a knowledge-based crossover is used. All of them used a random mutation operator which replaces a randomly selected candidate Web service with another random candidate Web service for the corresponding task.
\par
Lotfi and Mowlani~\cite{lotfi2011ica} utilized imperialist competitive algorithm~\cite{atashpaz2007imperialist} to obtain a near-optimal solution. Each solution is encoded as an integer vector. The assimilation operator used is based on a randomly created mask array. The revolution operator replaces a randomly selected candidate Web service with another random candidate Web service for the corresponding task.
\par
Wang, Sun, and Zou~\cite{wang2013particle} considered global QoS constraints and used a skyline operator~\cite{borzsony2001skyline} to prune redundant candidate Web services which cannot be part of the optimal solution in order to reduce the search space. Finally, by encoding each solution as an integer vector, particle swarm optimization~\cite{kennedy2010particle} is utilized to obtain a near-optimal solution.
\par
Wu and Zhu~\cite{wu2013transactional} considered transactional constraints and modeled the problem as a constrained directed acyclic graph. Finally, they utilized ant colony optimization~\cite{dorigo2010ant} to obtain a near-optimal solution.
\subsection{Non-Evolutionary Approaches}
Alrifai, Skoutas, and Risse~\cite{alrifai2010selecting} considered global QoS constraints and used a skyline operator to eliminate useless candidate Web services which cannot be part of the optimal solution in order to reduce the search space. To cope with cases that the number of candidate Web services is still too large to be practically beneficial, a set of representative candidate Web services is selected using a hierarchical clustering mechanism for further reducing the search space. Finally, mixed integer linear programming~\cite{nemhauser1988integer} is utilized to obtain a near-optimal solution.
\par
Alrifai, Risse, and Nejdl~\cite{alrifai2012hybrid, alrifai2009combining} considered global QoS constraints and used a hybrid approach to obtain a near-optimal solution. The hybrid approach utilizes mixed integer linear programming to seek the optimal decomposition of the global QoS constraints into local QoS constraints, and uses them as thresholds to perform local selection for each task in a distributed fashion.
\\\par
In spite of all the works done on the Web service selection problem, there is still need for more research. Firstly, to the best of our knowledge, there is no approach which supports all the constraints. Secondly, more efficient and effective approaches are essential for real-time applications. Therefore, different from the existing works, in this paper all global QoS constraints, interservice constraints, and transactional constraints are considered simultaneously, and a more efficient and effective approach using the Pareto principle and genetic algorithm, namely GAP2WSS, is proposed.
\section{Problem Formulation}
In this section, the Web service selection problem is formulated by considering global QoS constraints, interservice constraints, and transactional constraints. And, a formal problem statement is presented at the end.
\subsection{Composition Patterns}
There are four primary composition patterns which can be used to compose various Web services and create a composite Web service. These composition patterns are:
\begin{IEEEitemize}
\item
Serial: a sequence of Web services $\{s_1, \dotsc, s_n\}$ are executed consecutively.
\item
Loop: a Web service $s$ is executed repeatedly up to a maximum of $it$ iterations.
\item
Parallel (and split/and join): a set of Web services $\{s_1, \dotsc, s_n\}$ are executed concurrently and merge synchronized.
\item
Switch (exclusive split/exclusive join): a Web service $s_i$ from a set of Web services $\{s_1, \dotsc, s_n\}$ is executed based on the outcome of the switch condition evaluation at runtime.
\end{IEEEitemize}
\par
Fig.~\ref{fig:workflow} presents the workflow of a composite Web service. As is shown in the figure, the workflow contains all the serial, loop, parallel, and switch composition patterns.
{\renewcommand{\arraystretch}{1.6}
\begin{table*}[!t]
\caption{Aggregation Functions for Different QoS Attributes~\cite{alrifai2012hybrid}}
\label{tab:alrifai2012hybrid}
\centering
\begin{tabular}{c||c|c|c|c}
\hline
& \textbf{Serial} & \textbf{Loop} & \textbf{Parallel} & \textbf{Switch} \\
\hline
\hline
\textbf{Response Time} & $\sum\limits_{i=1}^n q(s_i)$ & ${it} \times q(s)$ & $\max\limits_{i=1}^n \{q(s_i)\}$ & $\max\limits_{i=1}^n \{q(s_i)\}$ \\
\hline
\textbf{Price} & $\sum\limits_{i=1}^n q(s_i)$ & ${it} \times q(s)$ & $\sum\limits_{i=1}^n q(s_i)$ & $\max\limits_{i=1}^n \{q(s_i)\}$ \\
\hline
\textbf{Availability} & $\prod\limits_{i=1}^n q(s_i)$ & $\prod\limits_{i=1}^{it} q(s)$ & $\prod\limits_{i=1}^n q(s_i)$ & $\min\limits_{i=1}^n \{q(s_i)\}$ \\
\hline
\textbf{Throughput} & $\min\limits_{i=1}^n \{q(s_i)\}$ & $q(s)$ & $\min\limits_{i=1}^n \{q(s_i)\}$ & $\min\limits_{i=1}^n \{q(s_i)\}$ \\
\hline
\end{tabular}
\end{table*}}
\subsection{Composite Web Service}
Let $T = \{T_1, \dotsc, T_{n'}\}$ be the set of all available tasks, and $T_i = \{s_{i,1}, \dotsc, s_{i,{m_i}}\}$ be the set of all candidate Web services providing the functionality for the task $i$, where $n'$, $m_i$ represent the number of available tasks and the number of available candidate Web services for the task $i$, respectively.
\par
Now, the composition request is represented by the vector of $CR = (S_1, \dotsc, S_n)$, which denotes the user's functional requirements for the composite Web service, where $n$ represents the number of tasks in the composite Web service, $S_i \in T$.
\par
Finally, the composite Web service is represented by the vector of $CS = (s_1, \dotsc, s_n)$, which denotes the selected candidate Web services for each task of the composition request, where $s_i \in S_i$.
\subsection{QoS}
The QoS values of a candidate Web service $s$ are represented by the vector of $Q_s = (q_1(s), \dotsc, q_k(s))$, where $k$ represents the number of QoS attributes for each candidate Web service. In addition, the vector of $Q = (Q_{1,1}, \dotsc, Q_{1,m_1}, \dotsc, Q_{i,1}, \dotsc, Q_{i,m_i})$ is consisted of the QoS values of all the candidate Web services for all the tasks. Finally, the QoS values of a composite Web service $CS$ is represented by the vector of $Q_{CS} = (q'_1(CS), \dotsc, q'_k(CS))$, where $q'_r(CS) = F_r{}_{i=1}^n(q_r(s_i))$, and $F_r$ indicates the aggregation function for the $r$th QoS attribute.
\par
The aggregation function depends on the QoS attribute and the composition pattern used. Table~\ref{tab:alrifai2012hybrid} shows aggregation functions of four different QoS attributes. It should be noted that for a composite Web service consisting of multiple tasks and multiple kinds of composition patterns, the aggregation functions are applied recursively.
\par
QoS attributes' weights are represented by the vector of $W = (w_1, \dots, w_k)$, which denotes the user's preferences with respect to different QoS attributes, where $w_r \in [0, 1]$, $\sum_{r=1}^{k} w_r = 1$.
{\renewcommand{\arraystretch}{1.3}
\begin{table*}[!t]
\caption{Derivation Rules for Transactional Properties~\cite{wu2013transactional, el2010tqos}}
\label{tab:wu2013transactional}
\centering
\begin{tabular}{c||cccc|c|cccc|cccc}
\hline
& $\mathbf{\to p}$ & $\mathbf{\to c}$ & $\mathbf{\to r}$ & $\mathbf{\to cr}$ & $\mathbf{*}$ & $\mathbf{\oplus p}$ & $\mathbf{\oplus c}$ & $\mathbf{\oplus r}$ & $\mathbf{\oplus cr}$ & $\mathbf{\otimes p}$ & $\mathbf{\otimes c}$ & $\mathbf{\otimes r}$ & $\mathbf{\otimes cr}$ \\
\hline
\hline
$\mathbf{p}$ & $\mathrm{\tilde{a}}$ & $\mathrm{\tilde{a}}$ & $\mathrm{p}$ & $\mathrm{p}$ & $\mathrm{\tilde{a}}$ & $\mathrm{\tilde{a}}$ & $\mathrm{\tilde{a}}$ & $\mathrm{\tilde{a}}$ & $\mathrm{p}$ & $\mathrm{p}$ & $\mathrm{p}$ & $\mathrm{p}$ & $\mathrm{p}$\\
\hline
$\mathbf{c}$ & $\mathrm{p}$ & $\mathrm{c}$ & $\mathrm{p}$ & $\mathrm{c}$ & $\mathrm{c}$ & $\mathrm{\tilde{a}}$ & $\mathrm{c}$ & $\mathrm{\tilde{a}}$ & $\mathrm{c}$ & $\mathrm{p}$ & $\mathrm{c}$ & $\mathrm{p}$ & $\mathrm{c}$ \\
\hline
$\mathbf{r}$ & $\mathrm{\tilde{a}}$ & $\mathrm{\tilde{a}}$ & $\mathrm{r}$ & $\mathrm{r}$ & $\mathrm{r}$ & $\mathrm{\tilde{a}}$ & $\mathrm{\tilde{a}}$ & $\mathrm{r}$ & $\mathrm{r}$ & $\mathrm{p}$ & $\mathrm{p}$ & $\mathrm{r}$ & $\mathrm{r}$ \\
\hline
$\mathbf{cr}$ & $\mathrm{p}$ & $\mathrm{c}$ & $\mathrm{r}$ & $\mathrm{cr}$ & $\mathrm{cr}$ & $\mathrm{p}$ & $\mathrm{c}$ & $\mathrm{r}$ & $\mathrm{cr}$ & $\mathrm{p}$ & $\mathrm{c}$ & $\mathrm{r}$ & $\mathrm{cr}$ \\
\hline
\end{tabular}
\end{table*}}
\subsection{Constraints}
Global QoS constraints are user's requirements regarding the values of different QoS attributes for a composite Web service, typically specified in the service level agreement~\cite{keller2003wsla}. They are represented by the vector of $QC = (c_1, \dotsc, c_k)$, where $c_r$ represents the constraint on the value of the $r$th QoS attribute of the composite Web service.
\par
The set of QoS attributes is consisted of two subsets: positive and negative attributes. The value of the positive QoS attributes (e.g., availability, throughput) should be maximized, whereas the value of the negative QoS attributes (e.g., response time, price) should be minimized. The positive and negative QoS attributes are represented by $Q^+, Q^-$, respectively. Thus, for a QoS attribute $r \in Q^+$, the constraint is a lower bound constraint and for a QoS attribute $r \in Q^-$, the constraint is an upper bound constraint. In addition, $c_r = \varnothing$ indicates no constraint on the $r$th QoS attribute.
\par
Interservice constraints are mutual context-based restrictions, such as technological, business, and partnership limitations between different candidate Web services. The set of interservice constraints is consisted of two subsets: dependency and conflict constraints. The interservice dependencies and interservice conflicts are represented by the set of $DC = \{(i, p, j, q) :$~if the candidate Web service $p$ is used for the task $S_i$, then the candidate Web service $q$ should be used for the task $S_j\}$, and $CC = \{(i, p, j, q) :$~if the candidate Web service $p$ is used for the task $S_i$, then the candidate Web service $q$ should not be used for the task $S_j\}$, respectively.
\par
A composite Web service requires transactional properties to guarantee its failure atomicity during runtime. A failure-atomic Web service ensures to succeed or fail as a complete unit. In other words, when it fails, all its previously committed activities can be rolled back with no effect at all. There are four primary failure-atomic transactional properties~\cite{mehrotra1992transaction, li2007deriving}:
\begin{IEEEitemize}
\item
Compensatable: a Web service is compensatable if it provides a compensation mechanism to roll back its effects when it is completed successfully. A composite Web service is compensatable if all its Web services are compensatable.
\item
Retriable: a Web service is retriable if it can be invoked repeatedly until it is completed successfully. A composite Web service is retriable if all its Web services are retriable.
\item
Pivot: a Web service is pivot if it is neither compensatable nor retriable. This means that it does not ensure to execute successfully and once it completes successfully, its effects cannot be rolled back.
\item
Compensatable and Retriable: a Web service in this class is both compensatable and retriable.
\end{IEEEitemize}
\par
The transactional constraints are represented by $TC \in \mathcal{P}(TP)$, where $TP =  \{\mathrm{p, c, r, cr}\}$ is the set of transactional properties and $\mathrm{p, c, r, cr}$ indicate the pivot, compensatable, retriable, and, compensatable and retriable property, respectively. It should be noted that $TC = \varnothing$ indicates no constraint on the transactional property of the composite Web service. In addition, the transactional property of a candidate Web service $s$ is represented by $tp(s)$ and the transactional property of a composite Web service $CS$ is represented by $tp'(CS)$.
\par
The transactional property of a composite Web service depends on the transactional properties of its Web services and the composition patterns used. Table~\ref{tab:wu2013transactional} shows derivation rules for different transactional properties, where $\to$, $\ast$, $\oplus$, $\otimes$ represent the serial, loop, parallel, and switch composition pattern, respectively, and $\mathrm{\tilde{a}}$ denotes non-atomicity.  It should be noted that for a composite Web service consisting of multiple tasks and multiple kinds of composition patterns, the derivation rules are applied recursively.
\subsection{Utility}
In order to map the QoS values of a candidate Web service to a single real value, for the purpose of sorting and ranking, a utility function is used. The utility function is defined using the simple additive weighting~\cite{zeleny1982multiple} technique. It involves a scaling phase to allow a uniform measurement of the QoS values independent of their ranges, and a weighting phase to represent user's preferences with respect to different QoS attributes. The utility value of a candidate Web service $s \in S_i$ is computed as follows:
\begin{equation}
\begin{aligned}
\label{eq:utility}
U(s) &= \sum\limits_{r \in Q^+} \frac{q_r(s) - Q_{min}(i, r)}{Q_{max}(i, r) - Q_{min}(i, r)} \cdot w_r \\ &+ \sum\limits_{r \in Q^-} \frac{Q_{max}(i, r) - q_r(s)}{Q_{max}(i, r) - Q_{min}(i, r)} \cdot w_r
\end{aligned}
\end{equation}
with:
\begin{displaymath}
\begin{aligned}
Q_{min}(i, r) &= \min\limits_{\forall s \in S_i} \{q_r(s)\} \\
Q_{max}(i, r) &= \max\limits_{\forall s \in S_i} \{q_r(s)\}
\end{aligned}
\end{displaymath}
where $Q_{min}(i, r)$, $Q_{max}(i, r)$ represent the minimum and maximum value for the $r$th QoS attribute of the candidate Web services of the task $S_i$, respectively. Finally, the utility value of a composite Web service $CS = (s_1, \dotsc, s_n)$ with a composition request $CR = (S_1, \dotsc, S_n)$ is computed as follows:
\begin{displaymath}
\begin{aligned}
U'(CS) &= \sum\limits_{r \in Q^+} \frac{q'_r(CS) - Q'_{min}(r)}{Q'_{max}(r) - Q'_{min}(r)} \cdot w_r \\ &+ \sum\limits_{r \in Q^-} \frac{Q'_{max}(r) - q'_r(CS)}{Q'_{max}(r) - Q'_{min}(r)} \cdot w_r
\end{aligned}
\end{displaymath}
with:
\begin{displaymath}
\begin{aligned}
Q'_{min}(r) &= F_r{}_{i=1}^n (Q_{min}(i, r)) \\
Q'_{max}(r) &= F_r{}_{i=1}^n (Q_{max}(i, r))
\end{aligned}
\end{displaymath}
where $Q'_{min}(r)$, $Q'_{max}(r)$ represent the minimum and maximum possible value for the $r$th QoS attribute of the composite Web service $CS$, respectively.
\par
It should be noted that the candidate Web services with a higher utility value are superior to the candidate Web services of the corresponding task with a lower utility value.
\subsection{Problem Statement}
\begin{definition}[Feasible Composite Web Service]
Given an octuple $\langle T, CR, Q, W, QC, DC, CC, TC \rangle$, a composite Web service $CS = (s_1, \dotsc, s_n)$ is feasible, if and only if all the global QoS constraints, interservice constraints, and transactional constraints are satisfied. Formally:
\begin{IEEEenumerate}
\item
$\forall r \in Q^- : q'_r(CS) \leq c_r$ and $\forall r \in Q^+ : q'_r(CS) \geq c_r$,
\item
$\forall d = (i, p, j, q) \in DC : \text{if } s_i = p \text{ then } s_j = q$,
\item
$\forall c = (i, p, j, q) \in CC : \text{if } s_i = p \text{ then } s_j \neq q$, and
\item
$tp'(CS) \in TC$.
\end{IEEEenumerate}
\end{definition}
\begin{definition}[Optimal Composite Web Service]
Given an octuple $\langle T, CR, Q, W, QC, DC, CC, TC \rangle$, a composite Web service $CS$ is optimal, if and only if it is a feasible composite Web service, and its utility value $U'(CS)$ is maximized along all the possible combinations.
\end{definition}
Now, with the prerequisite backgrounds in mind, the Web service selection problem can be formally stated as follows:
\begin{definition}[The Web Service Selection Problem]
Given an octuple $\langle T, CR, Q, W, QC, DC, CC, TC \rangle$ as input, return an optimal composite Web service $CS$ as output.
\end{definition}
\section{The Proposed Approach: GAP2WSS}
Genetic algorithms are search heuristics which rely on the principles of evolution by means of natural selection, namely selection, crossover, and mutation. Since, the Web service selection problem is a constrained combinatorial optimization problem, genetic algorithms might be efficient and effective for solving it.
\par
In this section, a genetic algorithm adopted by the Pareto principle, namely GAP2WSS, is proposed for solving the Web service selection problem. The pseudocode of the proposed approach is presented in Fig.~\ref{fig:algorithm}.
\begin{figure}
\begin{algorithmic}
\STATE score and rank all the candidate Web services per each task;
\STATE generate the initial population using the top 20 percent candidate Web services of each task;
\WHILE{the termination condition is not satisfied}
	\STATE probabilistically select parents and apply the crossover operator on them to create the offsprings;
	\STATE probabilistically select parents and apply the mutation operator on them to create the mutants;
	\STATE probabilistically select individuals from the current population, offsprings, and mutants to create the new population;
\ENDWHILE
\RETURN the best individual of the last iteration.
\end{algorithmic}
\caption{The pseudocode of GAP2WSS}
\label{fig:algorithm}
\end{figure}
\subsection{The Basic Idea}
The Pareto principle (a.k.a. the 80-20 rule)~\cite{koch201180}, states that 20 percent of causes, inputs, or efforts normally lead to 80 percent of the results, outputs, or rewards. This principle has many validated cases in economics, business, society, etc. The most popular case suggests that 80 percent of the wealth is in the hands of 20 percent of the people.
\par
The Pareto principle can be discreetly used to decrease the search space extremely, and thereupon increase the efficiency and efficacy of the approach used for solving the Web service selection problem. In order to utilize this principle, a mechanism to score and rank different candidate Web services of different tasks is used. This mechanism consists of two phases: scoring and ranking.
\par
In the scoring phase, at first, the utility value of a candidate Web service $s \in S_i$ with respect to its QoS values is calculated by~\eqref{eq:utility}. Then, the utility value of the candidate Web service $s$ with respect to the global QoS constraints is calculated by:
\begin{displaymath}
UC(s) = C_{max} - C(s)
\end{displaymath}
where $C_{max}$, $C(s)$ represent the maximum possible number of global QoS constraints violations, and the number of global QoS constraints violations of the candidate Web service $s$, respectively. Next, the utility value of the candidate Web service $s$ with respect to the interservice constraints is calculated by:
\begin{displaymath}
UV(s) = V_{max} - V'(s)
\end{displaymath}
where $V_{max}$, $V'(s)$ represent the maximum possible number of interservice constraints violations, and the number of interservice constraints on the candidate Web service $s$, respectively. Finally, the utility value of the candidate Web service $s$ with respect to the transactional constraints is calculated by:
\begin{displaymath}
UT(s) =
\left \{
\begin{array}{ll}
3, & \text{if } tp(s) = \mathrm{cr} \\
2, & \text{if } tp(s) = \mathrm{c} \text{ or } \mathrm{r}\\
1, & \text{if } tp(s) = \mathrm{p} \text{.}
\end{array}
\right.
\end{displaymath}
\par
The above scores are extracted from~\cite{wu2013transactional}. As the authors stated, a compensatable and retriable candidate Web service is the most suitable to compose a failure-atomic composite Web service, whereas a pivot candidate Web service is the least.
\par
In the ranking phase, at first, the rank of the candidate Web service $s$ among all the candidate Web services of the task $i$ is determined. The rank of the candidate Web service $s$ with respect to the QoS values, global QoS constraints, interservice constraints, and transactional constraints utility value is represented by $RQ(s)$, $RC(s)$, $RV(s)$, and $RT(s)$, respectively.
\par
Eventually, the global rank of the candidate Web service $s$ is computed by:
\begin{equation}
\label{eq:rank}
R(s) = \frac{RQ(s)}{RQ_{max}(i)} + \frac{c \times RC(s)}{RC_{max}(i)} + \frac{v \times RV(s)}{RV_{max}(i)} + \frac{t \times RT(s)}{RT_{max}(i)}
\end{equation}
with:
\begin{displaymath}
\begin{aligned}
RQ_{max}(i) &= \max\limits_{\forall s \in S_i} \{RQ(s)\} \\
RC_{max}(i) &= \max\limits_{\forall s \in S_i} \{RC(s)\} \\
RV_{max}(i) &= \max\limits_{\forall s \in S_i} \{RV(s)\} \\
RT_{max}(i) &= \max\limits_{\forall s \in S_i} \{RT(s)\}
\end{aligned}
\end{displaymath}
where if global QoS constraints exist $c = 1$ and otherwise $c = 0$, if interservice constraints exist $v = 1$ and otherwise $v = 0$, and if transactional constraints exist $t = 1$ and otherwise $t = 0$.
\subsection{Initial Population}
Each individual in the population represents a composite Web service. An individual $X$ is encoded as the vector of $X = (x_1, \dotsc, x_n)$, where $x_i \in S_i$.
\par
In order to generate the initial population, at first, the global rank of all the candidate Web services for all the tasks is calculated using~\eqref{eq:rank}. Then, 20 percent of the candidate Web services with the best ranks are selected as candidate Web services of each task. Finally, the Web service selection problem is solved by focusing only on these 20 percent candidate Web services of each task. More specifically in the genetic algorithm, the genes of each individual is randomly selected only from these 20 percent candidate Web services of the corresponding tasks.
\par
Eventually, $n_{pop}$ individuals are generated as the initial population $P(0)$, where $P(t)$ represents the population at the time $t$.
\subsection{Fitness Function}
{\renewcommand{\arraystretch}{2.3}
\begin{table*}[!t]
\caption{Fitness Function}
\label{tab:fitnessFunction}
\centering
\begin{tabular}{ccc||c|c|c||c}
\hline
\multicolumn{3}{c||}{\multirow{2}{*}{\textbf{Constraints}}} & \multirow{2}{*}{\textbf{Function} ($\bm{F}$)} & \textbf{Function} & \textbf{Corresponding} & \multirow{2}{*}{\textbf{Fitness Function ($\bm{FF}$)}} \\
\multicolumn{3}{c||}{} & & \textbf{Range ($\bm{fr}$)} & \textbf{Range ($\bm{cr}$)} & \\
\hline
\hline
$\bm{\frac{C(X)}{C_{max}}=0}$ & $\bm{\frac{V(X)}{V_{max}}=0}$ & $\bm{T(X)=0}$ & $U'(X)$ & $[0,1]$ & $[0.75,1]$ & $\bm{\frac{3+U'(X)}{4}}$ \\
\hline
$\bm{\frac{C(X)}{C_{max}}=0}$ & $\bm{\frac{V(X)}{V_{max}}=0}$ & $\bm{T(X)=1}$ & $U'(X)$ & $[0,1]$ & $[0.5,0.75)$ & $\bm{\frac{2+U'(X)}{4}}$ \\
\hline
$\bm{\frac{C(X)}{C_{max}}=0}$ & $\bm{\frac{V(X)}{V_{max}} \in (0, 1]}$ & $\bm{T(X)=0}$ & $U'(X)-\frac{V(X)}{V_{max}}$ & $[-1,1]$ & $[0.5,0.75)$ & $\bm{\frac{5+U'(X)-\frac{V(X)}{V_{max}}}{8}}$ \\
\hline
$\bm{\frac{C(X)}{C_{max}}=0}$ & $\bm{\frac{V(X)}{V_{max}} \in (0, 1]}$ & $\bm{T(X)=1}$ & $U'(X)-\frac{V(X)}{V_{max}}$ & $[-1,1]$ & $[0.25,0.5)$ & $\bm{\frac{3+U'(X)-\frac{V(X)}{V_{max}}}{8}}$ \\
\hline
$\bm{\frac{C(X)}{C_{max}} \in (0, 1]}$ & $\bm{\frac{V(X)}{V_{max}}=0}$ & $\bm{T(X)=0}$ & $U'(X)-\frac{C(X)}{C_{max}}$ & $[-1,1]$ & $[0.5,0.75)$ & $\bm{\frac{5+U'(X)-\frac{C(X)}{C_{max}}}{8}}$ \\
\hline
$\bm{\frac{C(X)}{C_{max}} \in (0, 1]}$ & $\bm{\frac{V(X)}{V_{max}}=0}$ & $\bm{T(X)=1}$ & $U'(X)-\frac{C(X)}{C_{max}}$ & $[-1,1]$ & $[0.25,0.5)$ & $\bm{\frac{3+U'(X)-\frac{C(X)}{C_{max}}}{8}}$ \\
\hline
$\bm{\frac{C(X)}{C_{max}} \in (0, 1]}$ & $\bm{\frac{V(X)}{V_{max}} \in (0, 1]}$ & $\bm{T(X)=0}$ & $U'(X)-\frac{C(X)}{C_{max}}-\frac{V(X)}{V_{max}}$ & $[-2,1]$ & $[0.25,0.5)$ & $\bm{\frac{5+U'(X)-\frac{C(X)}{C_{max}}-\frac{V(X)}{V_{max}}}{12}}$ \\
\hline
$\bm{\frac{C(X)}{C_{max}} \in (0, 1]}$ & $\bm{\frac{V(X)}{V_{max}} \in (0, 1]}$ & $\bm{T(X)=1}$ & $U'(X)-\frac{C(X)}{C_{max}}-\frac{V(X)}{V_{max}}$ & $[-2,1]$ & $[0,0.25)$ & $\bm{\frac{2+U'(X)-\frac{C(X)}{C_{max}}-\frac{V(X)}{V_{max}}}{12}}$ \\
\hline
\end{tabular}
\end{table*}}
Selecting the candidate Web service with the highest utility value for each task, does not ensure that the specified constraints are satisfied nor the utility value of the composite Web service is maximized. Hence, a fitness function is used to enable comparison of different individuals with respect to their utility values and the degree of constraints satisfaction.
\par
A genetic algorithm can only be applied to unconstrained problems directly. Hence, to handle the constraints, a penalty mechanism which penalizes the individuals who violate any constraints is used in defining the fitness function. It should be noted that the infeasible individuals are not eliminated from the population, because they may have some schemata which is required to obtain the optimal solution.
\par
Since, the constraints satisfaction is prior to the utility value, in other words a higher utility value does not justify the constraints violation, two principles are considered to define the fitness function. First, all the feasible individuals should have a higher fitness value than any infeasible individual. Second, the more constraints violated by an infeasible individual, the more penalty it receives.
\par
To this aim, the fitness function range $[0, 1]$ is divided up into four equal ranges. If no constraints are violated the fitness value will be in the range of $[0.75, 1]$, if one kind of constraints is violated the fitness value will be in the range of $[0.5, 0.75)$, if two kinds of constraints are violated the fitness value will be in the range of $[0.25, 0.5)$, and if all three kinds of constraints are violated the fitness value will be in the range of $[0, 0.25)$.
\par
Then, a function $F$ is defined to compare individuals with the same kinds of constraint violations. Each function has its own range, which is mapped to the corresponding range to obtain the fitness function $FF$ by:
\begin{displaymath}
FF = cr_{min} + \frac{(F - fr_{min}) \cdot (cr_{max} - cr_{min})}{fr_{max} - fr_{min}}
\end{displaymath}
where $cr_{min}$, $cr_{max}$ represent the minimum and maximum value of the corresponding range, respectively, and $fr_{min}$, $fr_{max}$ represent the minimum and maximum value of the function's range, respectively.
\par
Finally, the resultant fitness function is presented in Table~\ref{tab:fitnessFunction}, where $C(X)$, $V(X)$ represent the number of global QoS constraints violations and the number of interservice constraints violations for the corresponding individual, respectively. In addition, $T(X) = 1$ represents the violation of transactional constraints, and $T(X) = 0$ represents otherwise.
\subsection{Crossover}
The mechanism used to select parents for the crossover is rank-based, in which the individuals with the better rank are more likely to be selected. The probability that an individual $i \in P(t)$ is selected as a parent for the crossover is determined by:
\begin{displaymath}
p_i = \frac{n_{pop} - r_i + 1}{\sum\limits_{i=1}^{n_{pop}} r_i}
\end{displaymath}
where $r_i$ represents the rank of the individual $i$ based on its fitness value.
\begin{figure}
\centering
\includegraphics[width=\columnwidth]{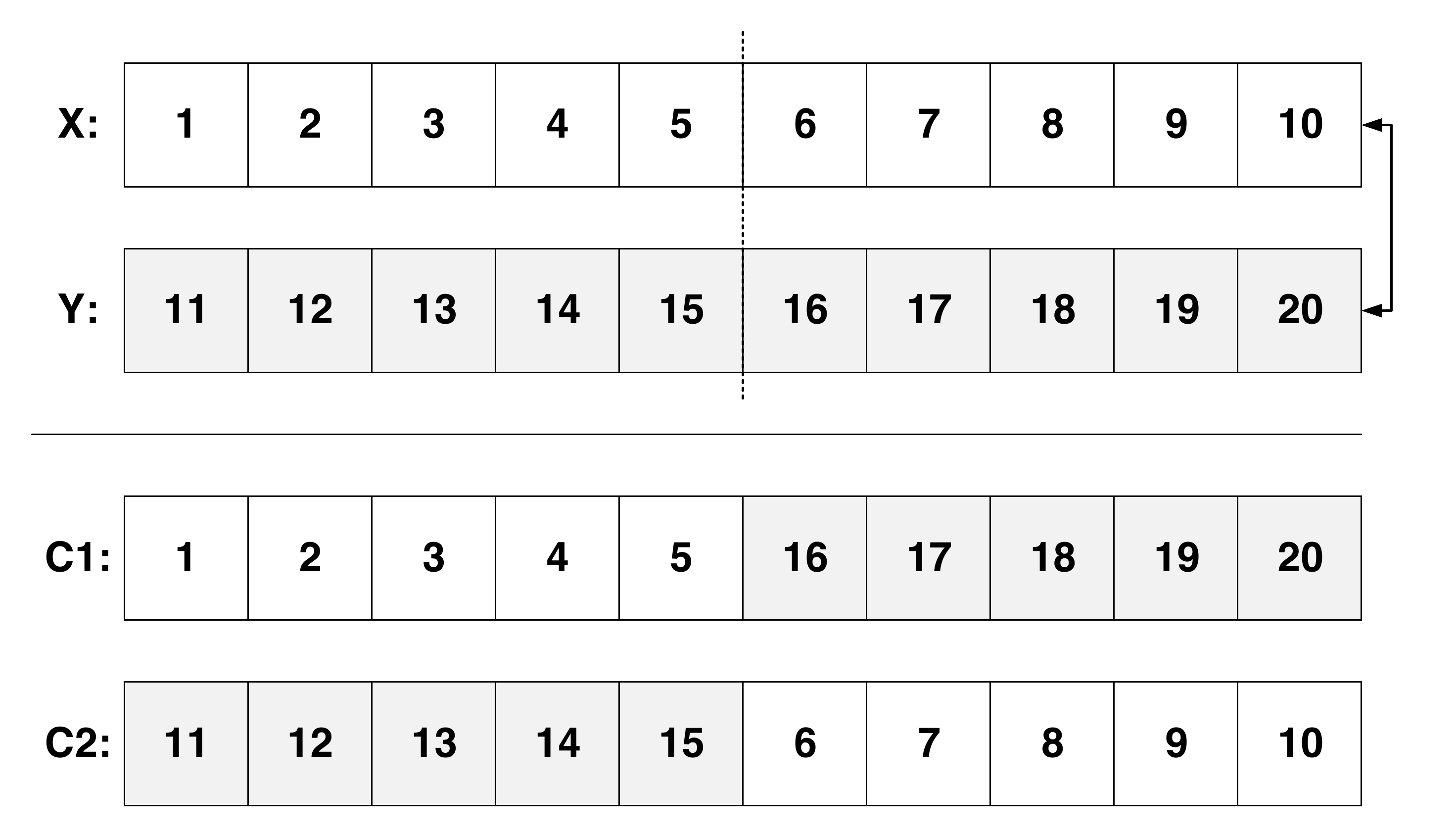}
\caption{Example of applying the crossover operator}
\label{fig:crossover}
\end{figure}
\par
The crossover operator used is a single-point crossover which takes two different individuals as parents and creates two different individuals as children. Let $X = (x_1, \dotsc, x_n)$ and $Y = (y_1, \dotsc, y_n)$ represent two parents and $l$ represent the crossover point. Then, two children $C_1$, $C_2$ are created as:
\begin{displaymath}
\begin{aligned}
C_1 &= (x_1, \dotsc, x_l, y_{l+1}, \dotsc, y_n) \\
C_2 &= (y_1, \dotsc, y_l, x_{l+1}, \dotsc, x_n).
\end{aligned}
\end{displaymath}
\par
Fig.~\ref{fig:crossover} demonstrates the crossover operator with an example. Eventually, $n_c = 2 \lceil \frac{p_c n_{pop}}{2} \rceil$ individuals are selected as parents and $n_c$ individuals are created as offsprings $Q(t)$, where $p_c$, $Q(t)$ represent the crossover rate and the offsprings at the time $t$, respectively.
\subsection{Mutation}
The mechanism used to select parents for the mutation is rank-based, in which the individuals with the better rank are more likely to be selected. The probability that an individual $i \in Q(t)$ is selected as a parent for the mutation is determined by:
\begin{displaymath}
p_i = \frac{n_{c} - r_i + 1}{\sum\limits_{i=1}^{n_{c}} r_i}
\end{displaymath}
where $r_i$ represents the rank of the individual $i$ based on its fitness value.
\begin{figure}
\centering
\includegraphics[width=\columnwidth]{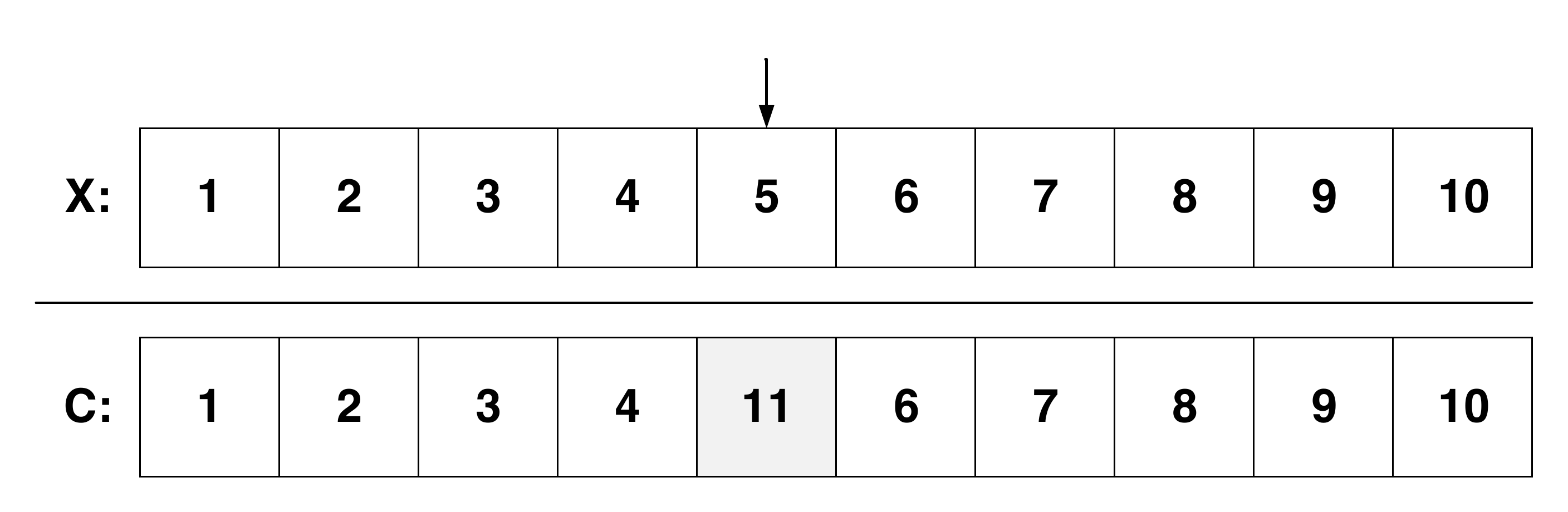}
\caption{Example of applying the mutation operator}
\label{fig:mutation}
\end{figure}
\par
The mutation operator used is a random mutation which takes an individual as a parent and creates an individual as a child. Let $X = (x_1, \dotsc, x_n)$ represent a parent and $l$ represent the mutation point. Then, $x_l$ is replaced by $x_l^{new} \in S_l - {x_l}$ to create a child $C$ as:
\begin{displaymath}
C = (x_1, \dotsc, x_l^{new}, \dotsc, x_n).
\end{displaymath}
\par
Fig.~\ref{fig:mutation} demonstrates the mutation operator with an example. Eventually, $n_m = \lceil p_m n_{pop} \rceil$ individuals are selected as parents and $n_m$ individuals are created as mutants $R(t)$, where $p_m$, $R(t)$ represent the mutation rate and the mutants at the time $t$, respectively.
\subsection{Replacement}
The mechanism used to select individuals of the new population is rank-based with elitism, in which the individuals with the better rank are more likely to be selected and the best individual is always transmitted to the new population. First, the offsprings $Q(t)$, the mutants $R(t)$ and the old population $P(t)$ are merged to create the temporary population $P'(t)$. Then, the probability that an individual $i \in P'(t)$ is selected as an individual of the new population is determined by:
\begin{displaymath}
p_i = \frac{n_{pop} + n_{c} + n_{m} - r_i + 1}{\sum\limits_{i=1}^{n_{pop} + n_{c} + n_{m}} r_i}
\end{displaymath}
where $r_i$ represents the rank of the individual $i$ based on its fitness value.
\par
Eventually, $n_{pop}$ individuals are selected to create the new population $P(t+1)$, which replaces the old population $P(t)$.
\section{Empirical Studies}
In order to investigate the efficiency and efficacy of GAP2WSS, extensive experiments are performed. In the following, after describing the experimental setup, the experiments are presented.
\subsection{Experimental Setup}
According to~\cite{tang2010hybrid}, the penalty-based genetic algorithm~\cite{ai2008penalty} is the fastest among the repairing-based genetic algorithm~\cite{ai2008qos} and the hybrid genetic algorithm~\cite{tang2010hybrid} for solving the Web service selection problem. For that reason, GAP2WSS is compared with the penalty-based genetic algorithm (PGA). Both GAP2WSS and PGA are implemented in Matlab R2014a, and all the experiments are conducted on a PC with Intel(R) Core(TM) i7 CPU at 3.40 GHz and 16 Gbytes RAM running 64-bit Windows 8.1 Enterprise.
\par
How to set the genetic algorithm parameters to enable the best performance is an optimization problem itself. Since, parameter optimization of genetic algorithms is problem-specific and is not the focus of this paper, similar to~\cite{tang2010hybrid}, the population size, crossover rate, and mutation rate of the algorithms are set to 100, 0.90, and 0.15, respectively. The termination condition of both the algorithms is \textit{exceeding a certain number of fitness function evaluations}. Since, there is no priori knowledge of the number of fitness function evaluations required for solving different test problems, GAP2WSS is first executed with a termination condition of \textit{no improvement on the best solution in 15 consecutive iterations} for each test problem. Next, the obtained number of fitness function evaluations is used as the termination condition of both the algorithms for the corresponding test problem. Since, unlike PGA, all global QoS constraints, interservice constraints, and transactional constraints are considered in GAP2WSS, the fitness function presented in this paper is used as the fitness function of both the algorithms. Moreover, due to the stochastic nature of genetic algorithms, both algorithms are executed 30 times for each test problem, and the average obtained fitness value of each test problem is recorded.
\par
For the purpose of evaluation, the workflow shown in Fig.~\ref{fig:workflow} is considered as the composite Web service. As is shown in the figure, the composite Web service is comprised of 10 tasks. To generate a composite Web service with more number of tasks, a certain number of the workflow is concatenated together. For example, constructing a composite Web service with 50 tasks is made by concatenating five copies of the workflow. After generating the tasks, the candidate Web services of each task are randomly selected from the QWS dataset. In addition, the transactional property of each candidate Web service is randomly assigned to one of the transactional properties from the set of $TP$. For simplicity, the weights of different QoS attributes are set to the same value. At last, the number of iterations of the loop in the workflow is fixed to five.
\\\par
The efficiency and efficacy of different approaches is dependent on the size and the complexity of the Web service selection problem. The size of the problem is a function of two factors: the number of tasks and the number of candidate Web services per each task. Besides, the complexity of the problem is determined by the number of global QoS constraints, interservice constraints, and transactional constraints. Therefore, five sets of test problems are generated to evaluate the impact of these factors independently. In the following, first, the experiments on the convergence and stability of the algorithms, and then, the experiments on each of these sets are presented.
\subsection{Experiments on the Convergence and Stability}
In this test, the number of tasks, candidate Web services per each task, global QoS constraints, interservice constraints, and transactional constraints is fixed to 10, 100, 1, 500, 4, respectively. In addition, the termination condition of the algorithms is \textit{250 iterations}.
\par
\begin{figure}[!t]
\centering
\includegraphics[width=\columnwidth]{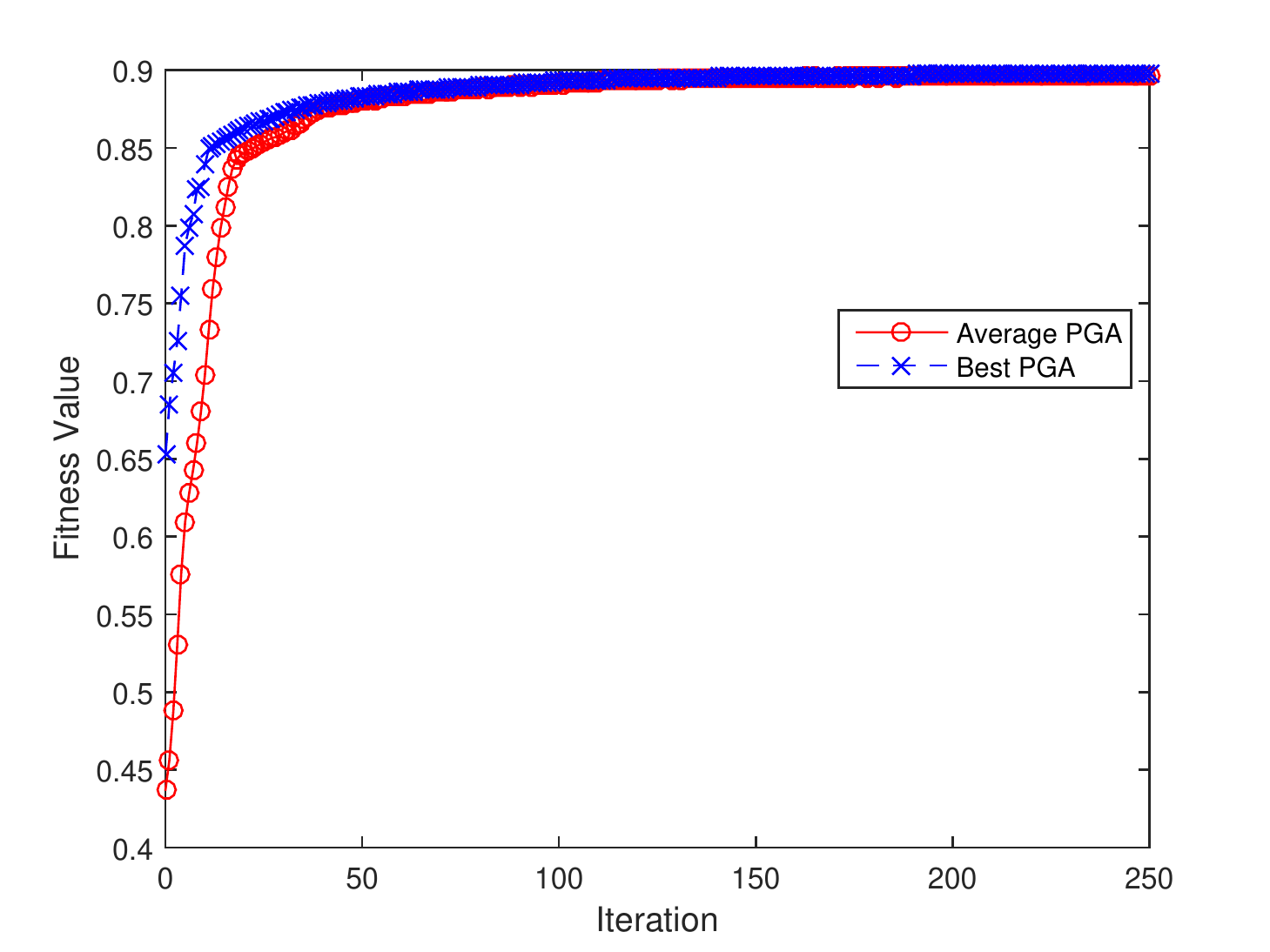}
\caption{The convergence diagram of PGA}
\label{fig:convergencePGA}
\end{figure}
\begin{figure}[!t]
\centering
\includegraphics[width=\columnwidth]{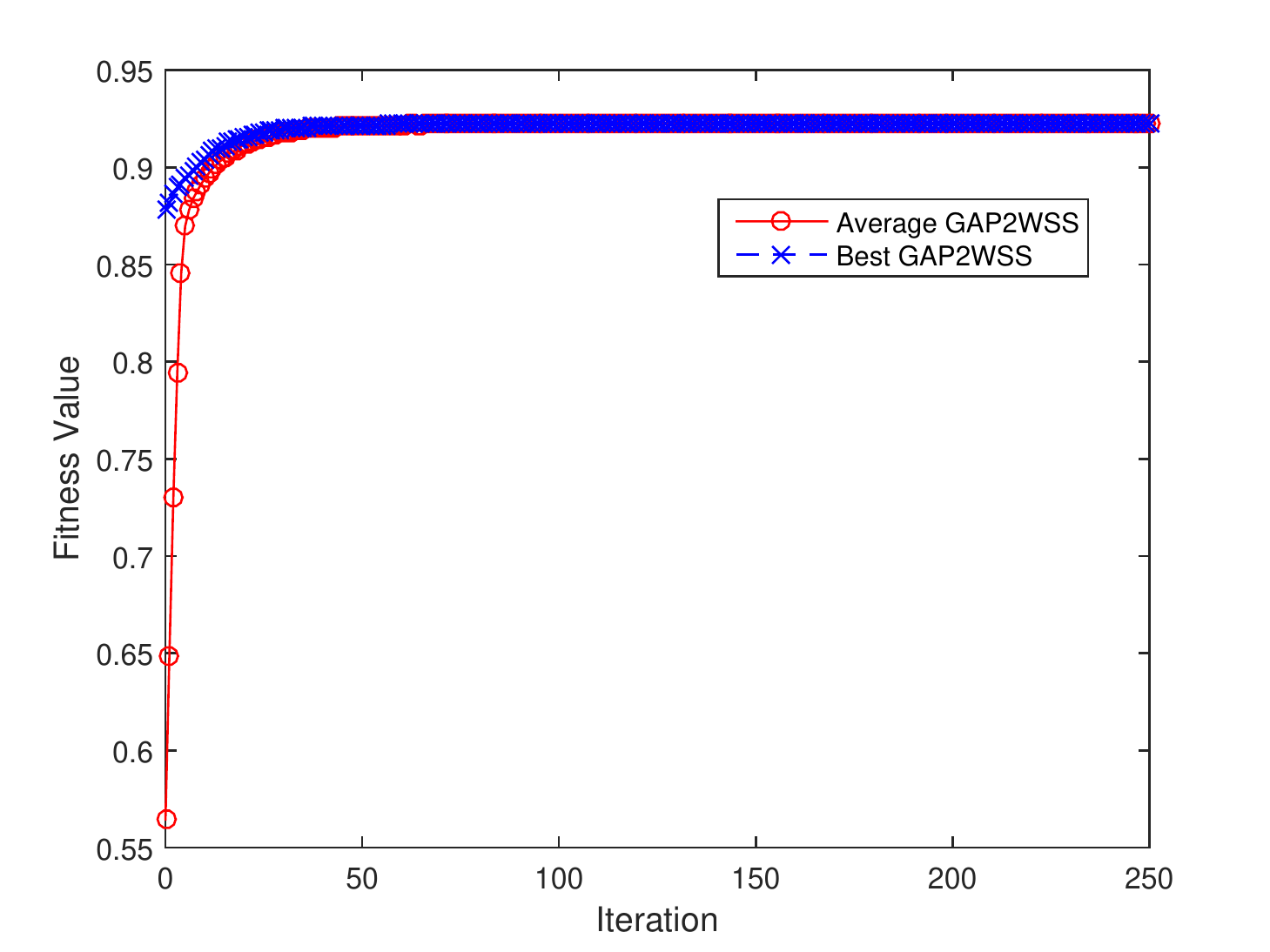}
\caption{The convergence diagram of GAP2WSS}
\label{fig:convergenceGAP2}
\end{figure}
\begin{figure}[!t]
\centering
\includegraphics[width=\columnwidth]{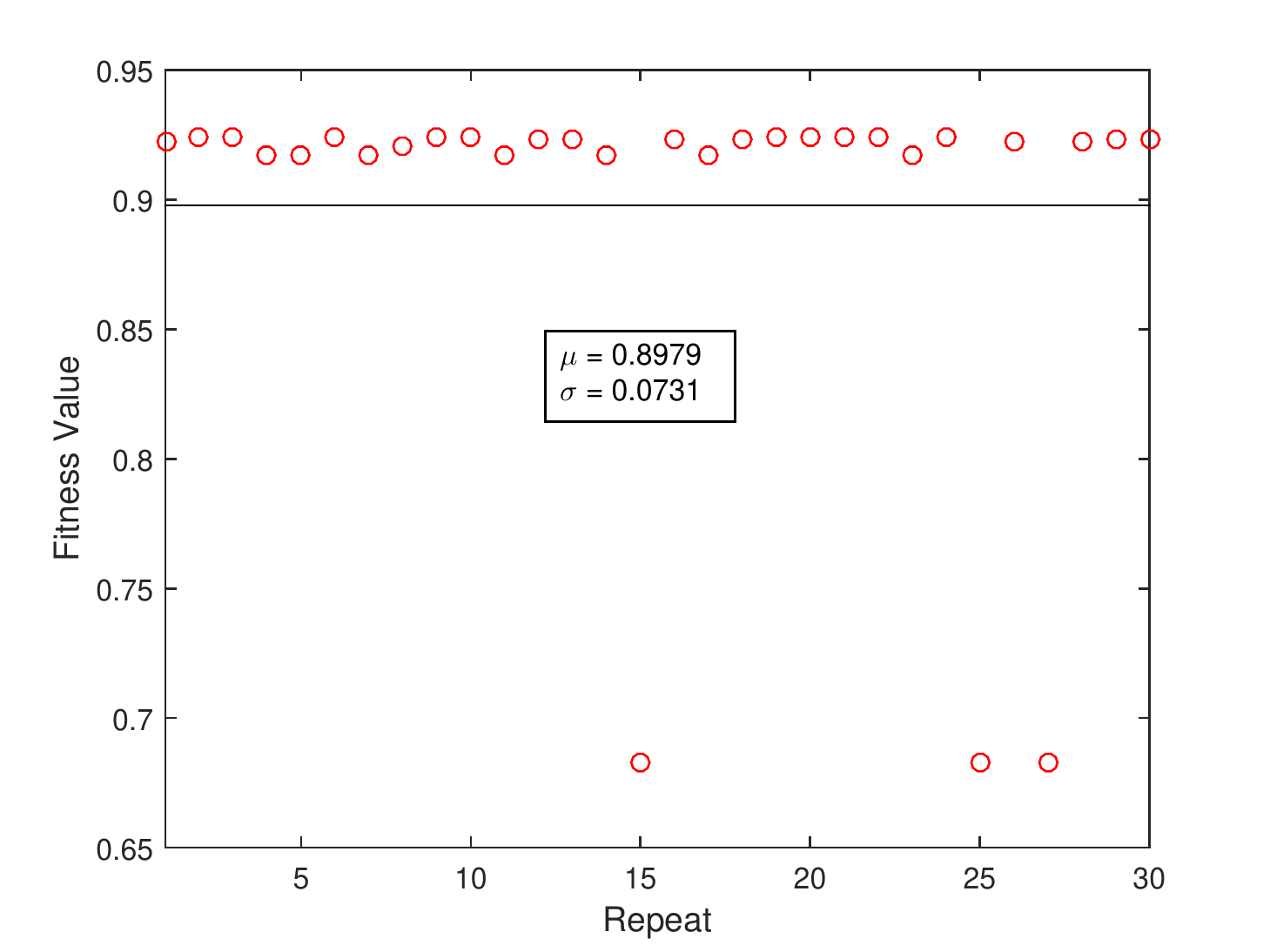}
\caption{The stability diagram of PGA}
\label{fig:stabilityPGA}
\end{figure}
\begin{figure}[!t]
\centering
\includegraphics[width=\columnwidth]{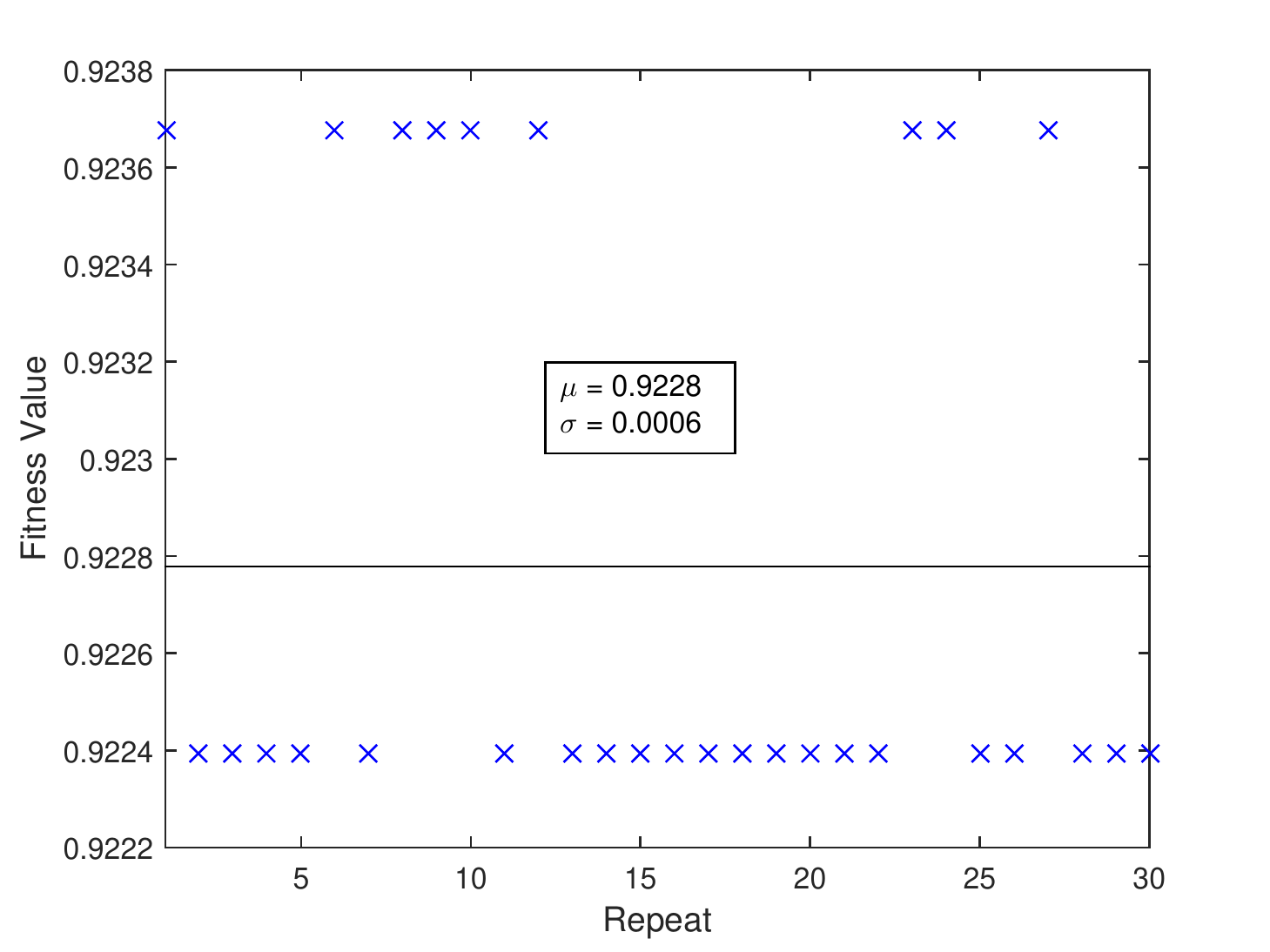}
\caption{The stability diagram of GAP2WSS}
\label{fig:stabilityGAP2}
\end{figure}

Fig.~\ref{fig:convergencePGA} and Fig.~\ref{fig:convergenceGAP2} depict the convergence diagram of PGA and GAP2WSS, respectively. In addition, the stability diagram of PGA and GAP2WSS is shown in Fig.~\ref{fig:stabilityPGA} and Fig.~\ref{fig:stabilityGAP2}, respectively. It can be noticed that GAP2WSS has a faster convergence and a higher stability than PGA.
\subsection{Experiments on the Number of Tasks}
In this set of test problems, the number of tasks is varied from 10 to 100 with an increment step of 10. In addition, the number of candidate Web services per each task is fixed to 500. In order to make sure that the results of the experiments are not biased by the constraints, the number of global QoS constraints, interservice constraints and transactional constraints are all fixed to zero.
\par
\begin{figure}[!t]
\centering
\includegraphics[width=\columnwidth]{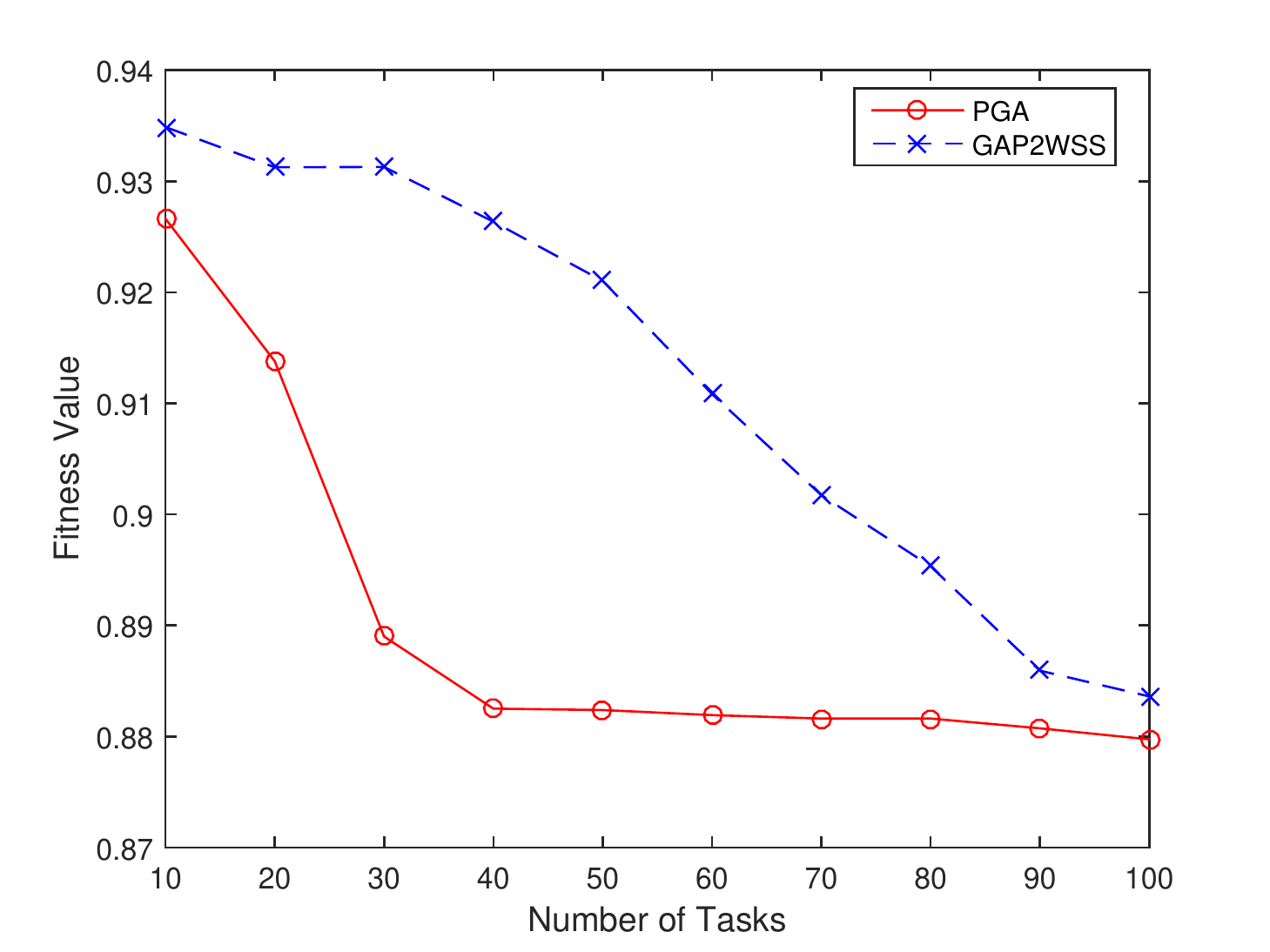}
\caption{Comparison of the approaches in terms of Fitness Value with respect to different Number of Tasks}
\label{fig:n}
\end{figure}

Fig.~\ref{fig:n} shows the fitness values obtained by the algorithms for the 10 test problems of this set. It can be seen that in all the test problems, GAP2WSS obtained higher fitness values than PGA. The average fitness value of this set of test problems is 0.9122, 0.8900 for GAP2WSS and PGA, respectively. In other words, the average fitness value of GAP2WSS to the average fitness value of PGA, has improved 2.49 percent for this set of test problems.
\subsection{Experiments on the Number of Candidate Web Services}
In this set of test problems, the number of candidate Web services per each task is varied from 100 to 1000 with an increment step of 100. In addition, the number of tasks is fixed to 50. In order to make sure that the results of the experiments are not biased by the constraints, the number of global QoS constraints, interservice constraints and transactional constraints are all fixed to zero.
\par
\begin{figure}[!t]
\centering
\includegraphics[width=\columnwidth]{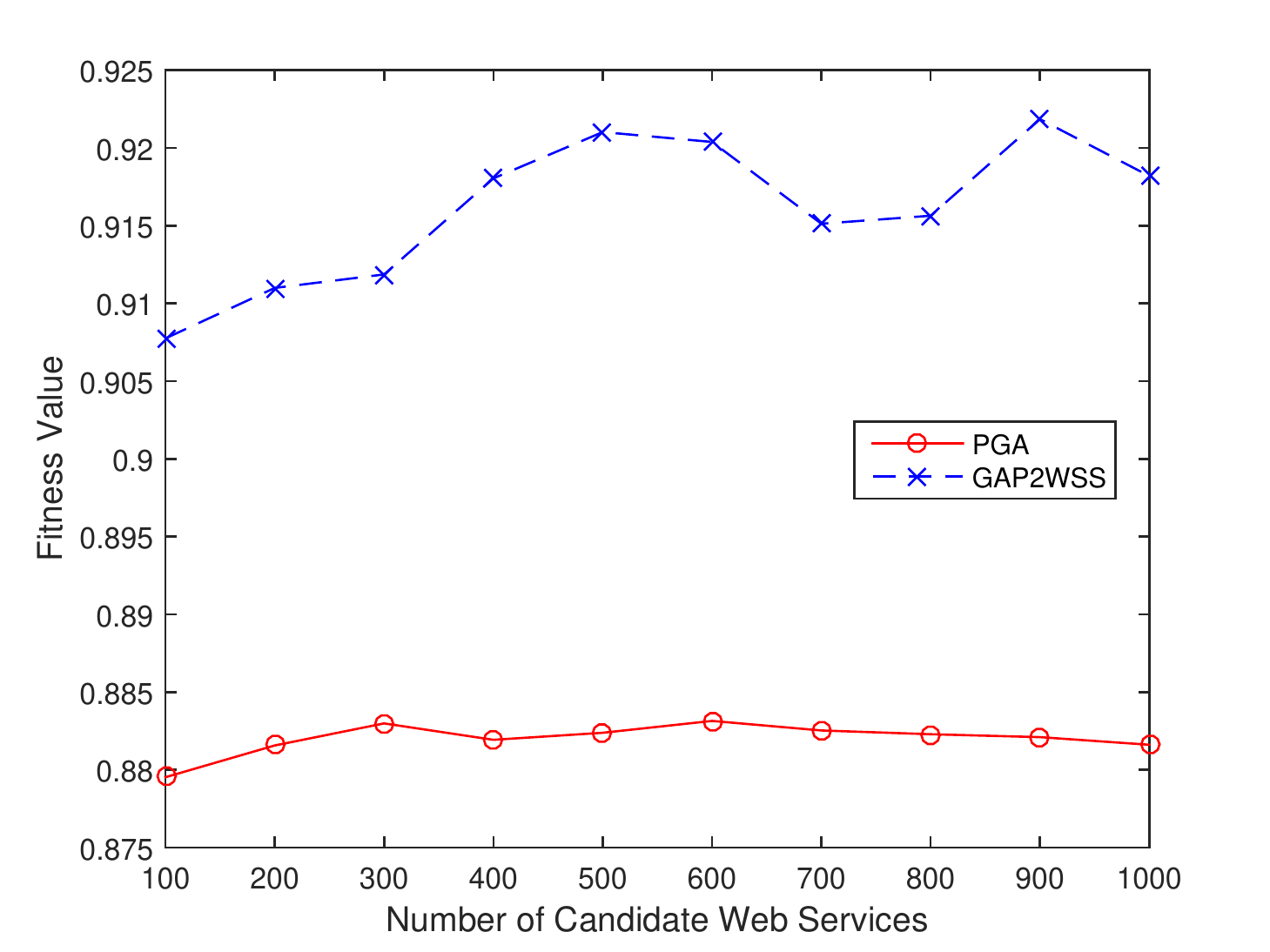}
\caption{Comparison of the approaches in terms of Fitness Value with respect to different Number of Candidate Web Services}
\label{fig:m}
\end{figure}

Fig.~\ref{fig:m} shows the fitness values obtained by the algorithms for the 10 test problems of this set. It can be noticed that in all the test problems, GAP2WSS obtained higher fitness values than PGA. The average fitness value of this set of test problems is 0.9161, 0.8820 for GAP2WSS and PGA, respectively. In other words, the average fitness value of GAP2WSS to the average fitness value of PGA, has improved 3.87 percent for this set of test problems.
\subsection{Experiments on the Number of Global QoS Constraints}
In this set of test problems, the number of global QoS constraints is varied from zero to nine with an increment step of one. In addition, the number of tasks and the number of candidate Web services per each task are fixed to 50 and 500, respectively. In order to make sure that the results of the experiments are not biased by the number of interservice constraints and transactional constraints, they are both fixed to zero. It should be noted that the average of minimum and maximum possible QoS values of the composite Web service are considered as the global QoS constraints.
\par
\begin{figure}[!t]
\centering
\includegraphics[width=\columnwidth]{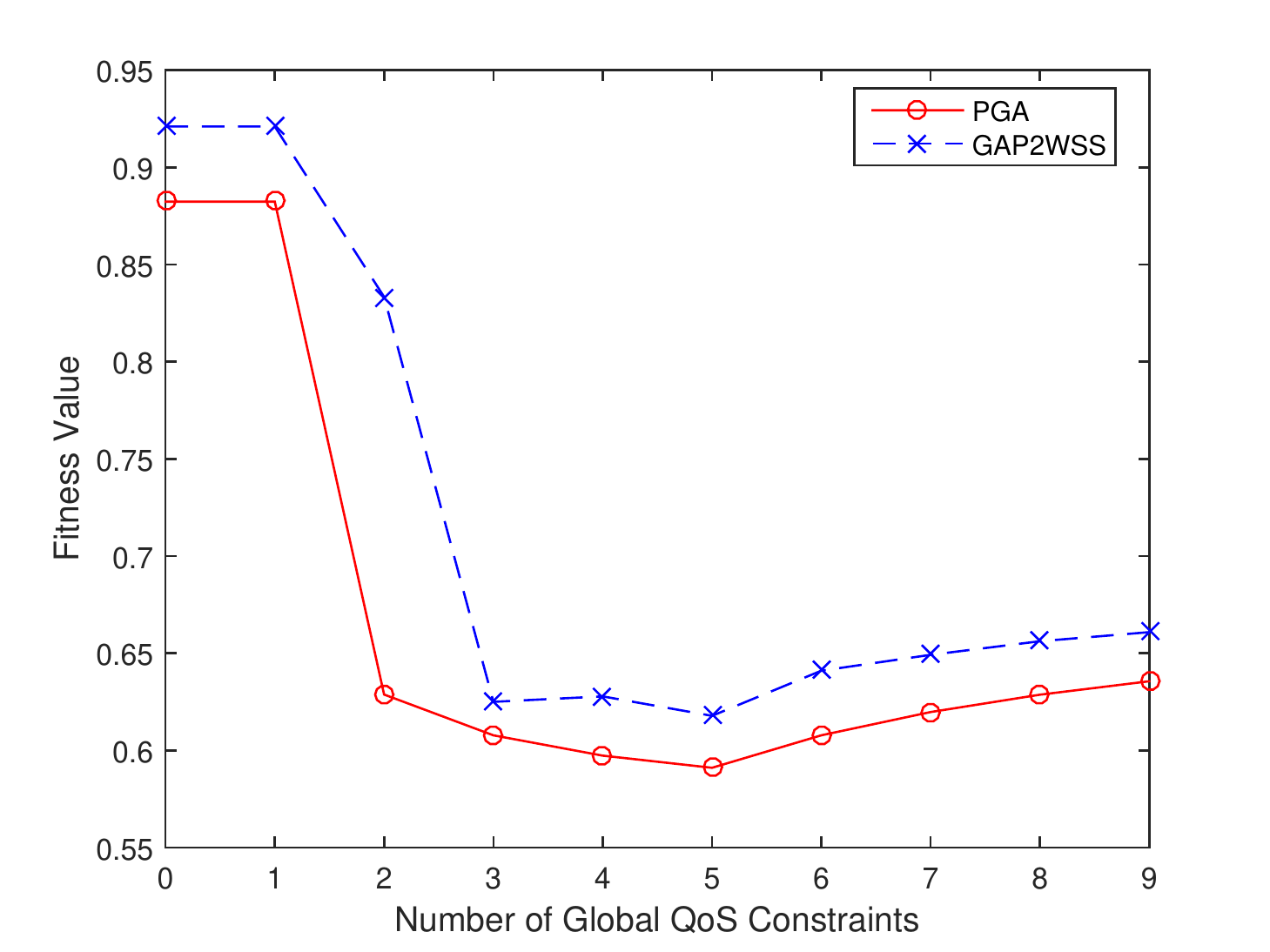}
\caption{Comparison of the approaches in terms of Fitness Value with respect to different Number of Global QoS Constraints}
\label{fig:qc}
\end{figure}

Fig.~\ref{fig:qc} shows the fitness values obtained by the algorithms for the 10 test problems of this set. It can be observed that in all the test problems, GAP2WSS obtained higher fitness values than PGA. The average fitness value of this set of test problems is 0.7153, 0.6681 for GAP2WSS and PGA, respectively. In other words, the average fitness value of GAP2WSS to the average fitness value of PGA, has improved 7.06 percent for this set of test problems.
\subsection{Experiments on the Number of Interservice Constraints}
In this set of test problems, the number of interservice constraints is varied from 0 to 5000 with an increment step of 500. In addition, the number of tasks and the number of candidate Web services per each task are fixed to 50 and 500, respectively. In order to make sure that the results of the experiments are not biased by the number of global QoS constraints and transactional constraints, they are both fixed to zero. It should be noted that the interservice constraints are generated randomly.
\par
\begin{figure}[!t]
\centering
\includegraphics[width=\columnwidth]{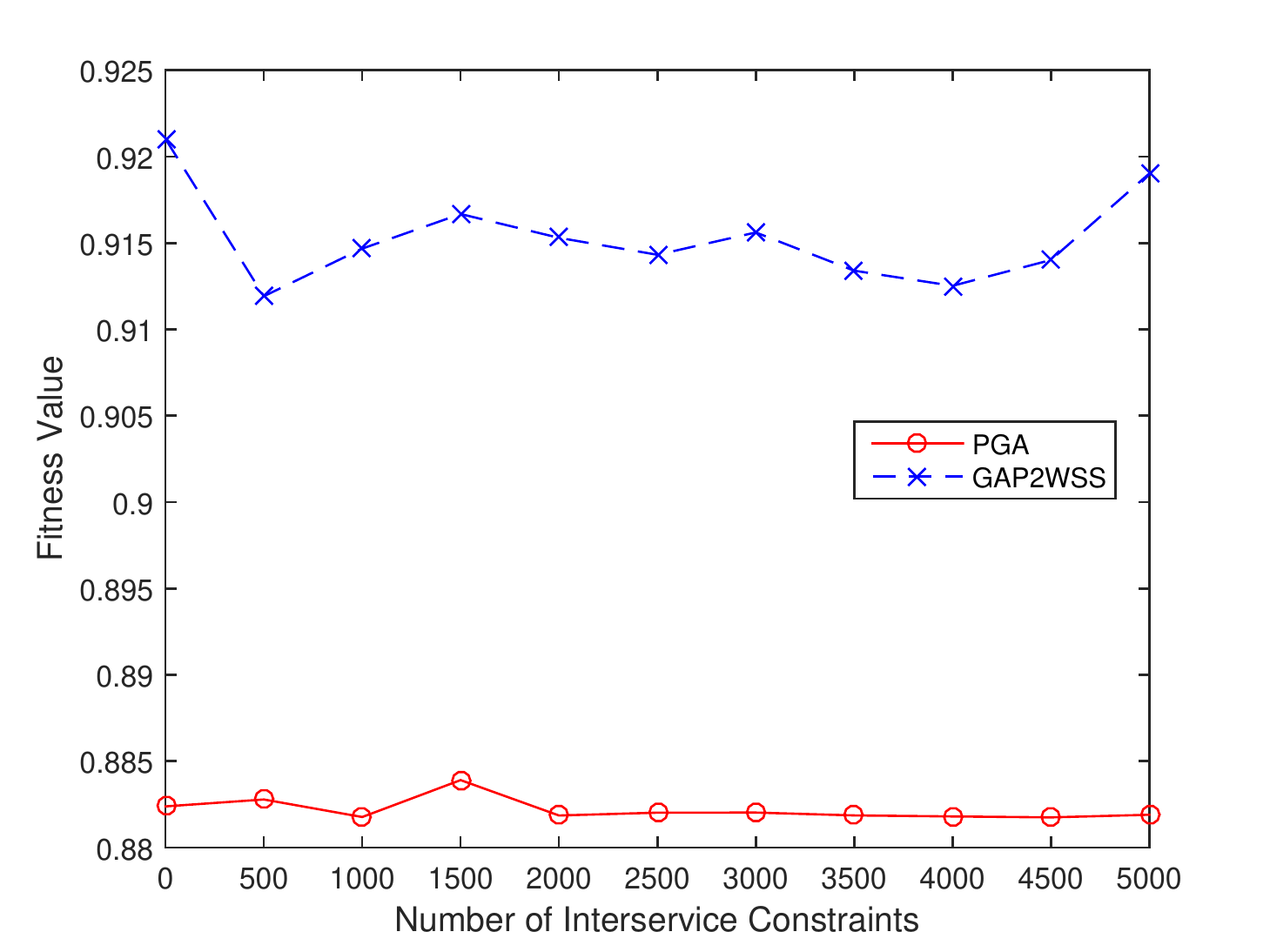}
\caption{Comparison of the approaches in terms of Fitness Value with respect to different Number of Interservice Constraints}
\label{fig:ic}
\end{figure}

Fig.~\ref{fig:ic} shows the fitness values obtained by the algorithms for the 11 test problems of this set. It can be seen that in all the test problems, GAP2WSS obtained higher fitness values than PGA. The average fitness value of this set of test problems is 0.9153, 0.8822 for GAP2WSS and PGA, respectively. In other words, the average fitness value of GAP2WSS to the average fitness value of PGA, has improved 3.75 percent for this set of test problems.
\subsection{Experiments on the Number of Transactional Constraints}
In this set of test problems, the number of transactional constraints is varied from zero to four with an increment step of one. In addition, the number of tasks and the number of candidate Web services per each task are fixed to 50 and 500, respectively. In order to make sure that the results of the experiments are not biased by the number of global QoS constraints and interservice constraints, they are both fixed to zero. It should be noted that the transactional constraints are picked from the set of $\mathcal{P}(TP)$.
\par
\begin{figure}[!t]
\centering
\includegraphics[width=\columnwidth]{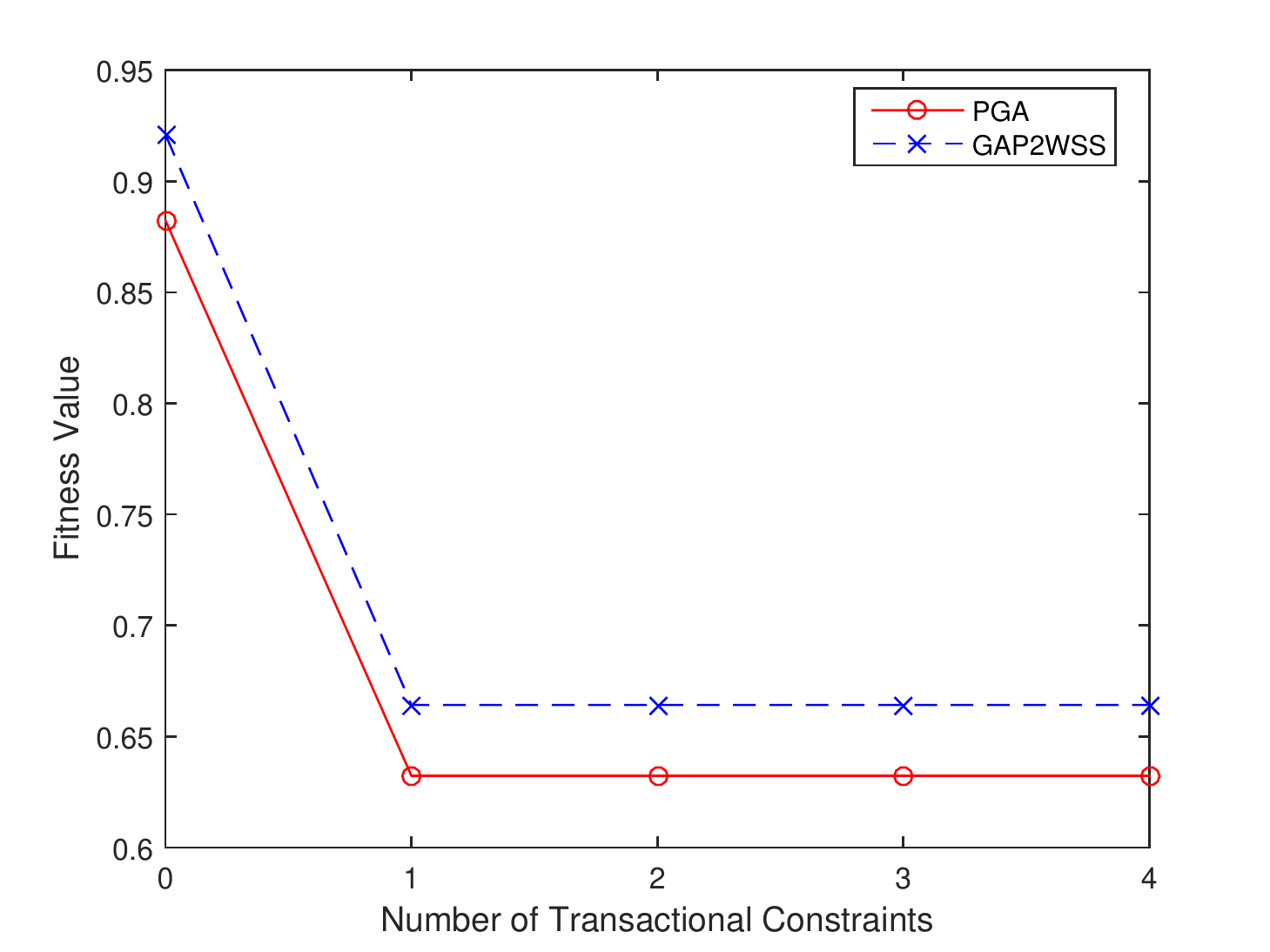}
\caption{Comparison of the approaches in terms of Fitness Value with respect to different Number of Transactional Constraints}
\label{fig:tc}
\end{figure}

Fig.~\ref{fig:tc} shows the fitness values obtained by the algorithms for the five test problems of this set. It can be noticed that in all the test problems, GAP2WSS obtained higher fitness values than PGA. The average fitness value of this set of test problems is 0.7155, 0.6823 for GAP2WSS and PGA, respectively. In other words, the average fitness value of GAP2WSS to the average fitness value of PGA, has improved 4.87 percent for this set of test problems.
\subsection{Discussion}
Clearly, it can be observed from the above experiments that GAP2WSS outperformed PGA in all the test problems, and handled different number of tasks, candidate Web services per each task, global QoS constraints, interservice constraints, and transactional constraints pretty good. The average fitness value of all the test problems is 0.8496, 0.8156 for GAP2WSS and PGA, respectively. In other words, the average fitness value of GAP2WSS to the average fitness value of PGA, has improved 4.17 percent for all the test problems.
\par
The main reason why GAP2WSS achieved far better fitness values than PGA is that the Pareto principle is adopted. This approach results in a reduced search space equal to $\sfrac{1}{5}$ of the original search space, which then leads to a quicker convergence to a reasonably good solution. Therefore, GAP2WSS has better efficiency and efficacy than PGA.
\par
In addition, it can be inferred that a near-optimal solution exists in this reduced search space, and there is no need to explore the remaining search space. Therefore, this mechanism can be used in other approaches for solving the Web service selection problem.
\section{Conclusion}
This paper presented a genetic algorithm based on the Pareto principle, namely GAP2WSS, for solving the Web service selection problem, which is known to be NP-hard in the strong sense. In the proposed approach, first, all the candidate Web services of different tasks are scored and ranked per each task. Then, using the Pareto principle, the problem search space is reduced to $\sfrac{1}{5}$ of the original search space. Finally, the problem is solved by focusing only on this reduced search space using a genetic algorithm.
\par
The efficiency and efficacy of the proposed approach is investigated by conducting extensive experiments. The results demonstrate that the proposed approach significantly increased the efficiency and efficacy and hence can be used in real-time applications. In addition, it can be inferred that a reasonably good solution exists in this reduced search space, and there is no need to explore the remaining search space. To the best of our knowledge, this is the first approach which supports all global QoS constraints, interservice constraints, and transactional constraints for solving the Web service selection problem.
\par
For future works, we suggest to propose a benchmark to enable fair comparison of different approaches for solving the Web service selection problem. By studying the advantages and disadvantages of different approaches, a more efficient, effective, and comprehensive approach for solving the problem should be proposed. In addition, more criteria should be considered for enterprise application requirements. Finally, we suggest to extend the usage of the Pareto principle to other approaches for the Web service selection problem and other problems.
\bibliographystyle{IEEEtran}
\bibliography{references}
\begin{IEEEbiography}[{\includegraphics[width=1in,height=1.25in,clip,keepaspectratio]{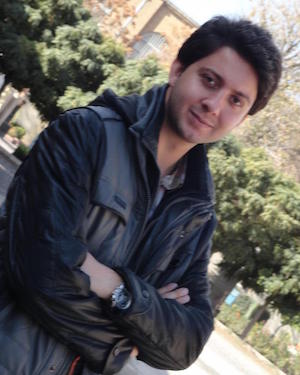}}]{SayedHassan Khatoonabadi}
received the BSc degree in computer science from Shahid Bahonar University of Kerman, Kerman, Iran, in 2012, and the MSc degree in computer science from University of Tabriz, Tabriz, Iran, in 2015. His research interests include service-oriented architecture, service-oriented computing, cloud computing, and Web services.
\end{IEEEbiography}
\begin{IEEEbiography}[{\includegraphics[width=1in,height=1.25in,clip,keepaspectratio]{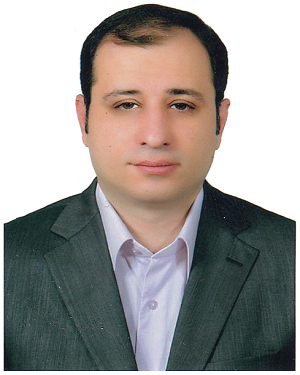}}]{Shahriar Lotfi}
received the BEng and MEng degrees in computer engineering all from University of Isfahan, Esfahan, Iran, in 1998 and 2002, respectively, and the PhD degree in computer engineering from Iran University of Science and Technology, Tehran, Iran, in 2008. Since 2008 he has been with the Department of Computer Science, Faculty of Mathematical Sciences, University of Tabriz, Tabriz, Iran, where he is currently an Assistant Professor, the Head of the department, and the Director of Intelligent Systems Laboratory. He has authored and co-authored over 30 publications. His research interests include parallel algorithms, supercompilers, and evolutionary algorithms.
\end{IEEEbiography}
\begin{IEEEbiography}[{\includegraphics[width=1in,height=1.25in,clip,keepaspectratio]{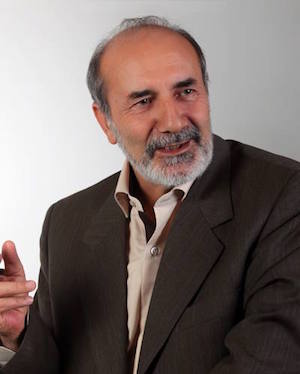}}]{Ayaz Isazadeh}
received the BSc degree in mathematics from University of Tabriz, Tabriz, Iran, in 1971, and the MSE degree in electrical engineering and computer science from Princeton University, NJ, US, in 1978, and the PhD degree in computing and information science from Queen's University,  ON, Canada, in 1996. From 1981 to 1990, he was a Member of Technical Staff in Bell Labs. In 1999, he founded and established the Department of Computer Science in Faculty of Mathematical Sciences, University of Tabriz, Tabriz, Iran, where he is currently a Full Professor and the Director of Computer Systems Laboratory. He was a Visiting Professor with the Department of Computer Engineering, Middle East Technical University Northern Cyprus Campus, Northern Cyprus, from 2012 to 2013. He has authored and co-authored over 90 publications. His research interests include automata theory and formal languages, software engineering, management information systems, information and communication technology, and, data structures and algorithms.
\end{IEEEbiography}
\end{document}